%% file: main.tex
\documentclass[letterpaper]{article} % DO NOT CHANGE THIS
\usepackage[T1]{fontenc}
\ifdefined\aaaianonymous
    \usepackage[submission]{aaai2026}  % Anonymous submission version
\else
    \usepackage{aaai2026}              % Camera-ready version
\fi

\usepackage{booktabs}
\usepackage{amsmath}
\usepackage{amsthm}
\usepackage{enumitem}
\usepackage{tabularx} 
\usepackage{mathtools}

\usepackage{color}
\usepackage{times}  % DO NOT CHANGE THIS
\usepackage{helvet}  % DO NOT CHANGE THIS
\usepackage{courier}  % DO NOT CHANGE THIS
\usepackage[hyphens]{url}  % DO NOT CHANGE THIS
\usepackage{graphicx} % DO NOT CHANGE THIS
\usepackage{natbib}  % DO NOT CHANGE THIS AND DO NOT ADD ANY OPTIONS TO IT
\usepackage{caption} % DO NOT CHANGE THIS AND DO NOT ADD ANY OPTIONS TO IT
\usepackage{algorithm}
\usepackage{algorithmic}

\usepackage{newfloat}
\usepackage{listings}
\DeclareCaptionStyle{ruled}{labelfont=normalfont,labelsep=colon,strut=off} % DO NOT CHANGE THIS
\floatstyle{ruled}
\newfloat{listing}{tb}{lst}{}
\floatname{listing}{Listing}

\nocopyright %-- Your paper will not be published if you use this command
\usepackage{amssymb} 
\usepackage{xcolor}

\title{Evaluating and Pricing Advertisements in AI-Generated Responses}

\author{
    John L. Turner-Smith\textsuperscript{\rm 1, \rm 2},
    Zimeng Huang\textsuperscript{\rm 2}\equalcontrib,
    Yuhan Fu\textsuperscript{\rm 2}\equalcontrib,
    Yihang Zhang\textsuperscript{\rm 2},
    Tonghan Wang\textsuperscript{\rm 2}
}
\affiliations{
    \textsuperscript{\rm 1}Carnegie Mellon University, Pittsburgh, PA, USA\\
    \textsuperscript{\rm 2}College of AI, Tsinghua University, Beijing, China
    jturners@andrew.cmu.edu,
    simona0212@foxmail.com,
    fuyh23@mails.tsinghua.edu.cn,
    hemmaxacand@gmail.com,
    twang1@g.harvard.edu

}

\usepackage{bibentry}

\begin{document}

\maketitle

\begin{abstract}
As search increasingly shifts toward LLM-driven answer engines, advertising is becoming embedded within the generated response itself and should therefore be evaluated for both user utility and commercial value. The key challenge is click-through intent: behavioural logs are unavailable, human annotation resists calibration, and frontier LLM judges conflate intent with linguistic fluency. These gaps compound, as principled pricing presupposes a continuous intent signal, while generating such a signal presupposes supervision that is currently unavailable. We construct the missing supervision through a psychologically grounded agent simulation framework, and distil it into a parameter-efficient evaluator that predicts click-through intent, together with the three companion dimensions of ad quality, as smooth, differentiable estimates. Validated through sign-certain behavioural perturbations, the evaluator surpasses frontier zero-shot judges on relevance sensitivity (79\% versus 60--67\%), tracks graded content degradation, generalises without error to 103 fictional products, and agrees with human preference in 86\% of pairwise judgements across five annotators, with agreement rising in the evaluator's confidence. Upon its estimates we build the pricing layer directly, deriving the unique payment rule under which truthful bidding is optimal, demonstrating it on a best-of-k allocation, and extending the mechanism to non-monotone allocations. The same differentiable signal stands ready as a training objective for ad generation.
\end{abstract}

\ifdefined\aaaianonymous
\else
\begin{links}
    % \link{Code}{https://aaai.org/example/code}
    % \link{Datasets}{https://aaai.org/example/datasets}
    % \link{Extended version}{https://aaai.org/example/extended-version}
\end{links}
\fi

\input{docs/1-intro}

\input{docs/2-related}

\input{docs/3-constructing}
\input{docs/4-evaluating}
\input{docs/5-auction}
\input{docs/6-discussion}
\input{docs/7-conclusion}

% ---------------------------------------------
% references
\bibliography{refs}
%----------------------------------------------
% ----------------------------------------------

\end{document}

%% file: docs/1-intro.tex
\section{Introduction}
\label{sec:intro}

\begin{figure*}[t]
\centering
\includegraphics[width=\textwidth]{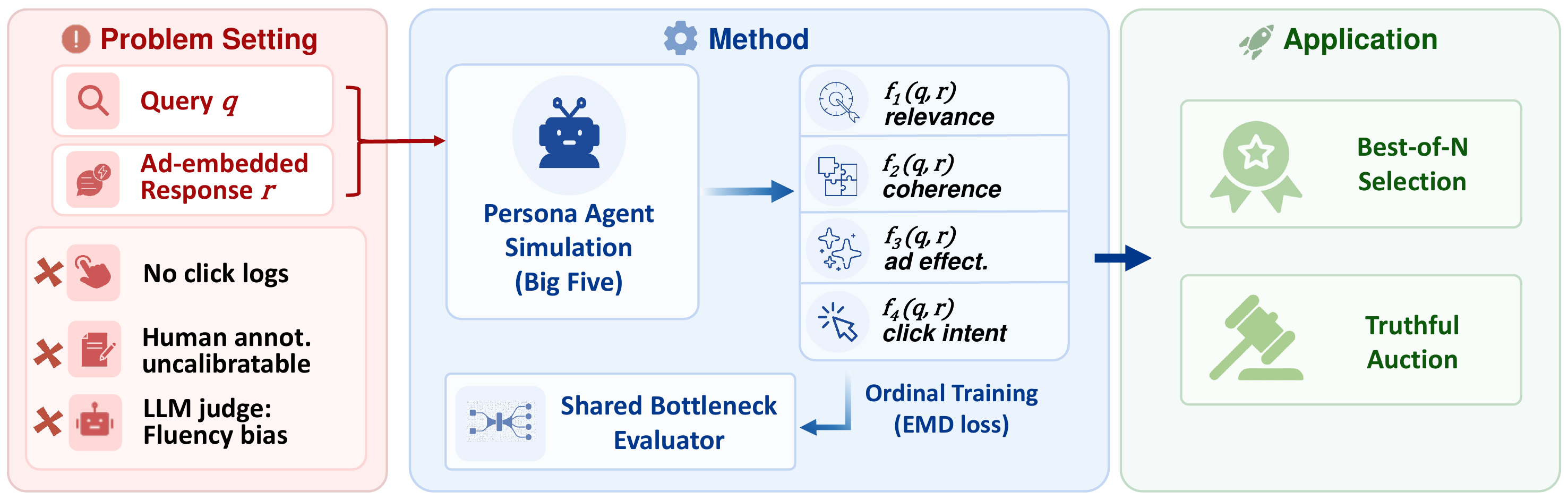}
\caption{Overview of the framework. From the absence of reliable click-intent supervision (left), a persona-agent simulation produces reproducible synthetic labels, distilled into a shared-bottleneck ordinal evaluator (centre) whose continuous output drives truthful auction pricing and Best-of-N ad selection (right).}
\label{fig:overview}
\end{figure*}

LLM-native advertising is fast becoming a principal monetisation channel for conversational AI. As users increasingly obtain information directly from LLM-powered answer engines, sponsored content is embedded within the generated response itself rather than presented alongside it. This shift creates a qualitatively different advertising format that must be evaluated along dual axes: user utility and commercial viability. A principled evaluation framework for such advertisements is therefore essential for generation quality, pricing, and user satisfaction.

Recent work has begun to address the evaluation of LLM-native advertising. NaiAD \cite{naiad} introduces a benchmark of 58,999 advertisement-embedded responses across four theoretically grounded dimensions spanning user and advertiser value. However, while its three semantic dimensions---response relevance, expression coherence, and advertising effectiveness---can be anchored and calibrated against human judgements, click-through intent (CTI) remains a big challenge to evaluate reliably. 

Three independent obstacles make reliable CTI evaluation intractable. First, unlike traditional display advertising, LLM-embedded ads in its nascent stage have generated no public historical click data. Second, human annotation is inherently uncalibratable; prior calibration efforts found that the null model was the most robust predictor for CTI, characterising it as a highly subjective variable that resists standard statistical calibration (Kendall's $W\!\approx\! 0.291$) \cite{naiad}. Third, prompting frontier LLMs as zero-shot judges yields a proxy for linguistic fluency, leaving the resulting score essentially unchanged (Section~\ref{sec:constructing}).

The inability to measure click-through intent paralyses the downstream commercial ecosystem. Principled incentive-compatible pricing~\cite{myerson1981optimal} demands a continuous, monotone mapping from bid to expected click probability, while generative ad optimisation requires a smooth, differentiable reward signal. Frontier LLMs, with their discrete and computationally costly text generation, are mathematically ill-suited to either role. Recent mechanism design for LLM advertising presupposes such an intent signal without constructing it in a log-free regime (see Section~\ref{sec:related}). This work aims to bridge the gap between presupposing this function and providing it.

We address this impasse through a unified framework connecting agentic simulation, neural evaluation, and commercial valuation. To overcome the absence of reliable supervision, we construct reproducible CTI labels using persona agents that decomposes click intent into objective features modulated by literature-grounded Big Five personality priors. We then train a continuous, differentiable evaluator on these synthetic labels using a shared-bottleneck architecture over a frozen Qwen3-4B backbone, optimised via an Earth Mover's Distance (EMD) objective to respect ordinal scale geometry. This architecture emerged from a comprehensive empirical investigation rather than a priori assumption: across the evaluated architectures, including flow-matching, diffusion-based, isolated-head, joint-and-isolated-head alternatives, the shared-bottleneck formulation was the only design to achieve stable and effective performance (Sections~\ref{subsec:architecture} and Section~\ref{subsec:arch_ablation}). Because no ground-truth target exists, we validate this evaluator intrinsically through rigorous, sign-certain behavioural perturbations and zero-shot semantic generalisation tests. Finally, we deploy the evaluator's continuous output to close the commercial loop, formally demonstrating that our intent signal directly enables Truthfulness-in-Expectation (TIE) mechanism design alongside Best-of-N generative ad optimisation.

The principal contributions of this work are as follows:
\begin{itemize}
\item \textbf{Reproducible Intent Supervision:} We introduce an agent-based labelling framework that constructs principled, reproducible click-through-intent supervision in a log-free, cold-start regime.
\item \textbf{A Differentiable, Intrinsically Validated Evaluator:} We develop a shared-bottleneck ordinal evaluator that passes a strict battery of sign-certain perturbation tests, outperforms zero-shot frontier judges on relevance sensitivity (79\% vs.\ 60--67\%), and generalises without error to 103 fictional products---evidence that it captures semantic intent rather than lexical memorisation.
\item \textbf{Closing the Economic and Generative Loop:} We carry the evaluator's output through to a truthful pricing mechanism, deriving the payment rule and demonstrating it on a best-of-$k$ allocation. In Best-of-N ad selection, the evaluator reaches 96\% agreement with majority human preference under high-confidence conditions.
\end{itemize}

%% file: docs/2-related.tex
\section{Related Work}
\label{sec:related}
\paragraph{LLM-native advertising and its evaluation.}
Sponsored content in language-model outputs has prompted conceptual analyses of the setting's opportunities and tensions \citep{feizi2023}, as well as datasets and benchmarks. NaiAD~\cite{naiad} provides advertisement-embedded responses and a four-dimensional evaluation framework with calibrated labels, flagging click-through intent as the dimension its calibration could not stabilise. GEM-Bench~\cite{gembench} offers a complementary benchmark spanning user satisfaction and engagement metrics. Work on ad insertion~\cite{xu2026} decouples placement from generation, finding a substantial fraction of readers do not recognise a sponsored recommendation even when disclosed~\cite{tang2025}. To our knowledge, no prior work directly addresses CTI evaluation in LLM-native advertising.

\paragraph{Click intent without behavioural logs.}
Classical click-through and engagement prediction \citep{rendle2010fm, cheng2016widedeep, guo2017deepfm, zhou2018din} learns from large logs of observed behaviour, which do not exist for advertisements embedded in language-model responses; recent cold-start work estimates propensity from advertisement content alone \citep{llmhyper}, though for display placements rather than generated prose. The directional effect of specific ad features is, by contrast, well established: personality-targeted appeals \citep{saha2024}, social proof \citep{bakshy2012}, and scarcity cues \citep{eisend2008} carry known, signed effects on engagement; these priors structure our feature decomposition and fix the direction of our sign-certain tests. Human annotation offers no escape, since disagreement on subjective judgements is often meaningful rather than incidental \citep{plank2022, davani2022} and learning under noisy labels remains hard \citep{song2022noisy}, motivating an agent-grounded alternative: language models can simulate human respondents \citep{argyle2023}, within documented limits \citep{sarstedt2024}, and normative Big Five distributions enable realistic persona sampling \citep{prosim}. Critically, prompted personality shapes a model's prose while leaving its numeric outputs largely unchanged \citep{psychmis2025}, motivating structural, feature-level weighting over prompt-based assignment (Section~\ref{sec:constructing}). Behavioural testing in NLP \citep{checklist} and metamorphic testing \citep{segura2016, chen2018metamorphic} require only that the direction of response to a controlled perturbation be known a priori.

\paragraph{Auctions and pricing for LLM advertising.}
A parallel literature develops auction and mechanism design for LLM advertising, encompassing token-level~\cite{dutting2024}, response-level~\cite{soumalias2024}, retrieval-augmented~\cite{hajiaghayi2024}, and summary-level~\cite{dubey2024} mechanisms; these works presuppose a continuous click or engagement signal without constructing it in a log-free regime.
The closest prior work to ours is \citet{llmauction}, which learns a click model as one component of an end-to-end generative auction; our evaluator differs in purpose and construction---a standalone, reusable module supervised by persona-agent labels and validated behaviourally, rather than a signal embedded within a single mechanism.

%% file: docs/3-constructing.tex
\section{Method: Click-Intent-Based Evaluation}
\label{sec:constructing}
As we have discussed in Section~\ref{sec:intro}, the absence of click logs and calibratable human annotations makes prompting a frontier LLM to estimate CTI from a given persona's perspective a plausible source of supervision. Our preliminary experiments, however, reveal a structural failure: the model’s prior preference for objectively high-quality text dominates its output, while realistic personality descriptions leaves final scores essentially unchanged (scores changes $|r| < 0.12$ given designated traits). Personality colours the generated prose but not the numerical prediction itself. We therefore generate reproducible synthetic labels through structured agentic simulation and distil this signal into a differentiable neural evaluator.

\textbf{Problem Formulation.} Let $q$ denote a user query and $r$ an ad-embedded response generated for it. The NaiAD framework assesses the pair $(q, r)$, hereafter an item, along four dimensions: response relevance (Q1), expression coherence (Q2), ad effectiveness (Q3), and click-through intent (Q4), each on a 1--5 scale. We seek an evaluator $f$ that maps $(q, r)$ to score estimates for all four dimensions. The downstream ecosystem imposes three requirements on the click-intent estimate $f_{\mathtt{CTI}}(q, r)$: it must be (1) deterministic, so that a response has a well-defined value; (2) continuous and differentiable, so that it be integrated into generative optimisation processes; (3) computationally efficient at inference, so that it can score candidate populations at scale. For pricing, let $G(v, q)$ denote a generation policy mapping an advertiser's bid $v$ and query $q$ to a response; the induced $x(v) \coloneqq \mathbb{E}_q[f_{\mathtt{CTI}}(G(v, q))] $ must then be monotone in $v$ for a truthful mechanism to exist (Section~\ref{sec:auction}). Q1--Q3 admit calibrated human supervision; Q4 does not \cite{naiad}. 

\subsection{Persona-Agent Labelling}
\label{subsec:persona_agent}
To prevent the model’s prior preference for objectively high-quality text from overriding personality traits, the persona-agent framework decouples evaluation into objective feature scoring and structured personality-based weighting.

\textbf{Stage 1: Objective Feature Assessment.} The agent first acts as a purely objective scorer, a task at which LLMs are highly consistent. For each advertisement, it evaluates six continuous features on a 1--5 scale: \textit{relevance}, \textit{copy quality}, \textit{novelty}, \textit{credibility}, \textit{urgency}, and \textit{warmth}. To denoise the single-call signal, each ad is assessed $K{=}3$ times and the scores are averaged. Inter-call reproducibility is high ($\sigma \approx 0.15$--$0.31$ across features), in contrast to the extreme variance of human CTI annotations ~\cite{naiad}.

\textbf{Stage 2: Structural Personality Weighting.}
Personality is introduced mathematically rather than linguistically.
We sample $K_\text{personas}=30$ personas from normative human population distributions of Big Five traits~\cite{prosim}.
Each trait is paired with the advertising feature it most plausibly amplifies, grounded in the personality-and-persuasion literature~\cite{hirsh2012,matz2017}:
\textit{Openness}\,$\to$\,novelty;
\textit{Extraversion}\,$\to$\,urgency;
\textit{Agreeableness}\,$\to$\,warmth;
\textit{Conscientiousness}\,$\to$\,credibility;
\textit{Neuroticism}\,$\to$\,credibility (as reassurance and security) and warmth at half weight.

Both the persona trait $z_t^{(p)}$ and the advertisement feature $z_f$ are z-centred against their population means, so that `high' denotes above-average, which matters because the raw features occupy mainly the low end of the scale.
The personality modulation for persona $p$ is:
\begin{equation}
m_p \;=\; \tanh\!\Big(\textstyle\sum_{(t,f)} s_{t,f}\,z^{(p)}_t\, z_f\Big),
\end{equation}
where the sum runs over the five trait--feature pairs and $s_{t,f}$ is the signed weight for each. Its sign is fixed by the direction of the trait's documented effect: positive where the trait amplifies receptivity to the paired feature, negative when it decreases receptivity of that feature.
The $\tanh$ bounds the personality effect to $(-1,1)$.
Per persona, the click-intent score is computed via a relevance gate:
\begin{align}
\text{base}_p &\;=\; \text{\emph{copy}}\cdot(\text{\emph{relevance}}/5), \\
c_p            &\;=\; \mathrm{clip}\!\big(\text{base}_p\cdot(1+\lambda\, m_p),\; 0.2,\; 5.0\big),
\end{align}
with $\lambda=0.5$ controlling the strength of the personality effect.
The gate ensures that an irrelevant advertisement cannot earn high click-through intent regardless of \emph{copy quality} or personality; personality enters only as a bounded modulation around a relevance-qualified base.
The final persona-agent label is $\bar{c} = \frac{1}{K_\text{personas}}\sum_p c_p$, the mean over all sampled personas.

\paragraph{Label properties.}
The label must discriminate, decompose non-redundantly, and weight its components as intended. It achieves a substantially wider dynamic range than a holistic baseline in which the agent scores click intent in a single call; each of the six features spans the full 1--5 range, and no feature pair exceeds $0.70$ correlation (achieved by sharpening \textit{novelty} to be independent of \emph{copy quality}). By construction, advertisement quality---particularly relevance---dominates, with personality a bounded modulation: the per-item spread across the 30 personas is small relative to, and coupled with, the item mean. The consequences of this relevance dominance are taken up in Section~\ref{sec:discussion}.

\subsection{Distilling a Differentiable Evaluator}
\label{subsec:architecture}
With reproducible intent labels constructed, we train a continuous, differentiable neural evaluator designed to serve as the unified scoring function for the downstream generative and economic ecosystem.
A shared bottleneck is preferred over per-dimension models because all four dimensions rest on the same language-comprehension substrate: joint training lets supervision from every dimension strengthen the shared representation, while dimension-specific heads preserve each dimension's calibration.

The evaluator is parameter-efficient: a frozen Qwen3-4B backbone adapted with rank-16 LoRA modules~\cite{lora}, whose pooled representation passes through a shared bottleneck before branching into four dimension-specific prediction heads; full architectural detail and hyperparameters are given in the supplementary material.

During training, the heads for Q1 (relevance), Q2 (coherence), and Q3 (ad effectiveness) are supervised by NaiAD's variance-calibrated human labels, while the click-through intent head is trained on our agent-grounded labels.
Because the agent-grounded labels are concentrated at the lower end of the scale, we parameterise the click-intent head's noise prior with their empirical mean and standard deviation rather than the NaiAD defaults---the labels themselves are unchanged---so the target distribution is represented faithfully during training. Training the CTI head jointly with the other three, rather than in isolation, sacrifices almost none of its accuracy---held-out Pearson 0.82 ($\sigma$ ratio 0.76) jointly, against 0.84 (0.83) for an isolated head---whilst yielding a single shared model.

\subsection{Ordinal Optimisation}
\label{subsec:ordinal_optimization}
Because the 1--5 evaluation scale possesses an inherent ordinal geometry, standard cross-entropy loss is  suboptimal. Instead, each prediction head is optimised using a squared Earth Mover's Distance (EMD~\cite{hou2016emd}) loss, which explicitly penalises predictions in proportion to their absolute ordinal distance from the target. The result is a direct score head that outputs a smooth, deterministic, and differentiable expected score.

%% file: docs/4-evaluating.tex
\section{Experiments}
\label{sec:evaluating}

In this section we evaluate the framework from three perspectives. We first report held-out fit across all four dimensions as the main result (Section~\ref{subsec:held_out_fit}). We then subject the evaluator to a battery of sign-certain behavioural perturbations, with zero-shot frontier judges as baselines on its two central tests, and test generalisation to wholly fictional products (Section~\ref{subsec:perturbations}). Finally, we anchor the signal in human judgement through a blind pairwise study of Best-of-N selection (Section~\ref{subsec:human_validation}).

\subsection{Experimental Setup}
\label{subsec:exp_setups}
\paragraph{Dataset and splits.} All experiments use NaiAD's 58,999 advertisement-embedded responses. The evaluator is trained and validated on a zero-overlap split with $n {=} 5{,}837$ held-out items; Q1--Q3 targets are NaiAD's variance-calibrated prediction-powered inference (VC-PPI) human labels, and the Q4 target is the agent-grounded label of Section~\ref{subsec:persona_agent}. The perturbation battery comprises 80 constructed items (15 cross-category swap pairs, six advertisements corrupted at five degradation levels, 10 keyword-stuffing pairs) and two diagnostics over fixed-seed samples of the labelled corpus ($n {=} 300$ for quintile calibration, $n {=} 200$ for input ablation); the generalisation test uses 103 fabricated product categories disjoint from the 1,348 product names in the training corpus.

\paragraph{Models and training.} The evaluator is the shared-bottleneck architecture of Section~\ref{sec:constructing}, trained with DeepSpeed ZeRO-2 in mixed precision under a cosine learning-rate schedule with warm-up, with separate learning rates for the LoRA modules and prediction heads; full hyperparameters are listed in the supplementary material.

\paragraph{Baselines.} We compare against three zero-shot frontier judges given identical inputs: Qwen3.6-35B-A3B (the evaluator's backbone family), Claude Sonnet 4.6, and GPT-5.5.

\paragraph{Evaluation metrics.} Held-out fit is reported as Pearson and Spearman correlation, RMSE, and the prediction-to-label standard deviation ratio $\sigma_{\text{pred}}/\sigma_{\text{label}}$; perturbation tests report directional pass rates against the a-priori-known sign; human validation reports evaluator-annotator agreement by confidence bin.

\paragraph{Architecture Ablation.} 
\label{subsec:arch_ablation}
We ablate whether modelling the full label distribution improves over direct ordinal regression. A flow-matching auxiliary head was co-trained; a controlled ablation revealed that disabling the flow-matching objective left the accuracy of all four dimensions entirely unchanged (click-intent Pearson: 0.821 with the auxiliary head versus 0.821 without). We therefore adopt the direct ordinal head alone, which is simpler at inference and yields the deterministic gradients that downstream pricing and Best-of-N selection require.

\subsection{Main Result: Held-out Fit Across All Four Dimensions}
\label{subsec:held_out_fit}
We first verify that the evaluator faithfully reproduces its supervision on the held-out split. For Q1--Q3, the targets are NaiAD's VC-PPI human labels; for click-through intent, the target is the agent-grounded label. The evaluator achieves strong Pearson correlations across the board: $0.85$ for Q1 (relevance), $0.73$ for Q2 (coherence), $0.55$ for Q3 (ad effectiveness), and $0.82$ for click-through intent; full per-dimension fit statistics are given in Table~\ref{tab:fit4d}.

\begin{table}[h]
\centering
\small
\renewcommand{\arraystretch}{1.2}
\begin{tabularx}{\columnwidth}{@{} X X cccc @{}}  % 前两列自动分配剩余宽度，后四列居中
\toprule
\textbf{Dim.} & \textbf{Target} & \textbf{Pearson} & \textbf{Spearman} & \textbf{RMSE} & \textbf{$\sigma_{\text{pred}}/\sigma_{\text{label}}$}\\
\midrule
Q1         & VC-PPI & 0.85 & 0.41 & 0.40 & 0.86 \\ % relevance
Q2        & VC-PPI & 0.73 & 0.57 & 0.50 & 0.72 \\ % coherence
Q3 & VC-PPI & 0.55 & 0.40 & 0.76 & 0.60 \\ % ad-effectiveness
Q4    & agent  & 0.82 & 0.73 & 0.44 & 0.76 \\ %  click-intent
\bottomrule
\end{tabularx}
\caption{Held-out fit per dimension ($n=5{,}837$). Q1--Q3 are evaluated against NaiAD's VC-PPI labels; Q4 against the agent-grounded label.}
\label{tab:fit4d}
\end{table}

Two apparent anomalies dissolve on inspection: Q1's low Spearman reflects label concentration (two-thirds of items fall within a 0.2-wide band, leaving little rank structure to recover), and Q3's weaker fit reflects compressed prediction range over the most widely spread labels.

\subsection{Sign-Certain Behavioural Perturbations}
\label{subsec:perturbations}
\begin{figure*}[t]
\centering
\includegraphics[width=\textwidth]{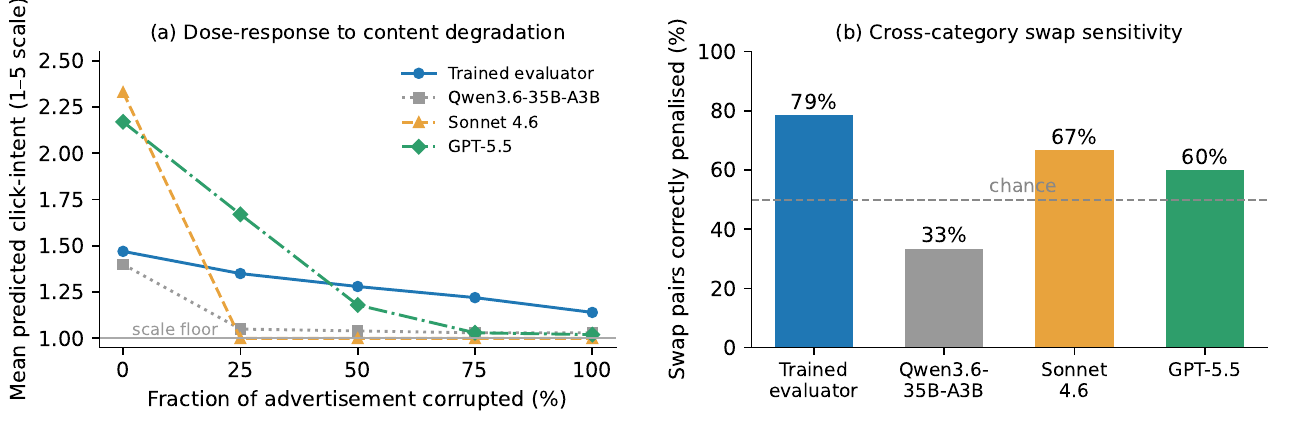}
\caption{Behavioural comparison with zero-shot judges. (a) Mean predicted click-intent as an increasing fraction of the advertisement is replaced with neutral filler; absolute values are not comparable across systems, so the relevant quantity is the shape of each curve. (b) Fraction of cross-category swap pairs on which each system correctly assigns lower click-intent to the foreign advertisement, tied pairs excluded from each system's  denominator; the dashed line marks chance.}
\label{fig:judges}
\end{figure*}

Because the exact magnitude of real-world click-intent is unknown, we validate the model using sign-certain behavioural perturbations---interventions where the correct direction (sign) of the score shift is known \textit{a priori}. A valid evaluator must lower its score when content degrades, and resist superficial statistical gaming. We subjected the evaluator to six stringent tests; on the two central ones---cross-category swap and graded degradation---we additionally administered the identical test items to the three zero-shot frontier judges of Section~\ref{subsec:exp_setups} as baselines, testing the natural objection that a state-of-the-art LLM, suitably prompted, could deliver the same signal zero-shot.
\begin{enumerate}
    \item \textbf{Cross-category swap:} Replacing the original ad with a foreign, off-category product lowers the predicted click-intent in 79\% of cases (11/14; one tied pair excluded), well above chance. The same-backbone zero-shot judge fails entirely (33\%, below chance), whilst the frontier judges Sonnet 4.6 and GPT-5.5 reach only 67\% and 60\% (Figure~\ref{fig:judges}b), differing by a single pair and hence not meaningfully ranked against each other.
    \item \textbf{Dose-response to content degradation:} As an increasing fraction of the ad is replaced with neutral filler, the predicted click-intent decreases monotonically across five corruption levels, dropping smoothly from 1.47 (unaltered) to 1.14 (fully corrupted) on the 1--5 scale. The zero-shot judges act instead as rigid classifiers: the Qwen judge and Sonnet 4.6 collapse to their minimum at the first corruption increment, and GPT-5.5 grades the lighter levels before hitting an abrupt floor (Figure~\ref{fig:judges}a). Only the trained evaluator yields the smooth, monotone gradient that downstream pricing and generative optimisation require, and it does so in a single, fast, inexpensive forward pass.
    \item \textbf{Keyword stuffing:} Padding the ad with repeated product names and aggressive calls to action lowers the score, suggesting the model has learned copy quality rather than rewarding superficial token density.
    \item \textbf{Quintile calibration:} Binning predictions into quintiles reveals that the mean true label rises monotonically across bins, ascending smoothly from 0.79 to 2.39.
    \item \textbf{Input ablation:} Blanking the ad shifts the prediction comparably to blanking the query, confirming the model fuses both inputs.
    \item \textbf{Fictional-product generalisation:} An evaluator trained on a fixed corpus of 1,348 product names might in principle owe its relevance sensitivity to memorised lexical co-occurrence rather than semantic understanding. To rule this out, for each of 103 fabricated product categories, none of whose names appear in the training corpus, one contextually relevant and one irrelevant placement of the same advertisement were generated. The evaluator assigns the relevant placement the higher click-intent score in all 103 cases (relevant mean 3.70, irrelevant mean 1.26; mean gap 2.44 on the 1--5 scale, smallest gap exceeding a full scale point). With zero token overlap with training, the discrimination is necessarily semantic.
\end{enumerate}
Together these tests constitute the primary evidence that the evaluator has internalised a meaningful intent signal rather than a surface correlate.

\subsection{Human Validation: Best-of-N and Pairwise Agreement}
\label{subsec:human_validation}
Finally, we validate the framework's downstream utility by closing the generative loop.
The evaluator's continuous output serves as a deterministic reward signal for Best-of-N (BoN) selection: given 15 candidate responses generated per query by Qwen3.6-35B-A3B, the evaluator selects the highest-scoring placement.
We validate these selections through a forced-choice, blind pairwise human evaluation: five annotators were each presented with 100 pairs of generated advertisements and asked which was more likely to be clicked, with the evaluator's predicted scores withheld (inter-annotator agreement: Fleiss $\kappa {=} 0.64$).
Majority human preference agrees with the evaluator's preference in 92\% of pairs (mean per-annotator agreement 86\%), and agreement rises monotonically with the evaluator's confidence gap, reaching 96\% when the predicted score gap exceeds 1.5, validating the evaluator as a reliable gatekeeper for automated generative ad pipelines.

%% file: docs/5-auction.tex
\section{From Intent Signal to Truthful Auction}
\label{sec:auction}

The evaluator of the preceding sections estimates, for a given response, how likely a user is to engage with its embedded advertisement. This estimate is the missing input to a question the evaluation of an advertisement must ultimately answer: what is that placement worth? We show that the click-intent output supplies exactly the primitive a principled pricing mechanism requires, derive the payment rule that follows, and demonstrate it upon an allocation grounded in the evaluator's own judgements. Full proofs are provided in the supplementary material.

\textbf{The Pricing Gap in LLM-Native Advertising.} 
Traditional search auctions price discrete slots against historical click-through rates. LLM-native advertising has neither: the generated response is the entire slot, and expected click probability varies continuously with how far the advertiser's intent shapes the prose. A truthful auction therefore requires an allocation function $x(v)$ mapping bid to expected engagement, a function that has hitherto lacked a principled foundation.

\textbf{Truthfulness-in-Expectation (TIE) Enabled by the Evaluator.} 
Our evaluator supplies this missing primitive as the empirical realisation of $x(v)$. Formally, let $G(v,q)$ denote the platform's generation policy that maps bid $v$ and query $q$ to an ad-embedded response; we define $x(v):=\mathbb{E}_{q}[f(G(v,q))]$, where $f$ is the evaluator's click-intent output. The output is a measurable, deterministic mapping from generated text to intent, and on it we build a single-parameter mechanism $(x, p)$ satisfying Truthfulness-in-Expectation (TIE), Individual Rationality (IR), and No Positive Transfers (NPT); its differentiability, though not required by the characterisation, is what additionally suits it as a training signal for generation (Section~\ref{sec:discussion}).

By the characterisation we derive (Theorem~A.2 of the supplementary material), a single-parameter mechanism is TIE if and only if $x(v)$ is monotonically non-decreasing in $v$.
Rather than assume the platform's policy induces a monotone allocation---an assumption a learned policy need not satisfy---we construct one. We instantiate the bid as optimisation effort: a bid of $k$ purchases $k$ candidate responses; that is, the platform converts the declared value into optimisation effort at a unit rate. It scores each candidate with the evaluator and returns the best. The induced allocation is the expected maximum of the click-intent estimate over the $k$ draws,
\begin{equation}
x(k) \;=\; \mathbb{E}\!\left[\max_{1\le j\le k} f(r_j)\right],
\end{equation}
which is non-decreasing in $k$ of necessity, the maximum over more draws never falling. Monotonicity here is a property of the construction rather than a hope about the policy.
Since the allocation is monotone by construction, the payment rule $p(v)$ is uniquely determined up to the normalisation constant $U(0)\ge 0$ by the payment identity (Theorem~A.2).
Under the canonical normalisation $U(0)=0$---which additionally enforces NPT, ruling out platform subsidies---the expected payment reduces to:
\begin{equation}
p(v) = v\,x(v) - \int_0^v x(t)\,dt
\end{equation}
Because advertisers in this ecosystem are ultimately charged on a per-click basis rather than an expected-impression basis, we derive the per-click price $\pi(v)$ by substituting the payment identity into $\pi(v)=p(v)/x(v)$, giving the explicit form (Proposition~A.3):
\begin{equation}
\pi(v) = v - \frac{1}{x(v)}\int_0^v x(t)\,dt
\end{equation}
The per-click IR bounds $0\le\pi(v)\le v$ follow from monotonicity of $x$ (Proposition~A.4); no advertiser therefore pays more per click than their declared valuation.

\textbf{A Worked Example.}
For a single advertiser and query, we draw a population of $k{=}30$ candidate responses from the PILA generator~\cite{pila} and score each with the evaluator. The evaluator reports its estimate on a five-point scale, which we carry into the unit interval by the order-preserving transform $(f-1)/4$; this preserves the ranking of responses but is not calibrated to absolute click frequencies, a limitation inherited from Section~\ref{sec:discussion}, so the example illustrates the mechanism's structure rather than predicting realised revenue. When the bid takes integer values, we apply the identity to the induced step allocation (with $x(0){=}0$, a null bid drawing nothing), giving $p(k){=}k\,x(k) - \sum_{j=1}^{k} x(j)$; truthfulness, individual rationality, and no-positive-transfers follow from Theorem~A.2 exactly as in the continuous case, and the continuum form is recovered as the grid refines. Table~\ref{tab:pricing} reports the schedule. The guarantees hold exactly upon this learned allocation: the allocation rises with the bid; the payment rises with it from $p(1){=}0$; and the per-click price respects $0\le\pi(k)\le k$ throughout (Proposition~A.4).

\begin{table}[t]
\centering
\small
\begin{tabular}{@{}rcccc@{}}
\toprule
bid $k$ & $\bar f_k$ & $x(k)$ & $p(k)$ & $\pi(k)$ \\
\midrule
1  & 3.48 & 0.621 & 0.000 & 0.000 \\
2  & 3.59 & 0.647 & 0.026 & 0.040 \\
3  & 3.63 & 0.657 & 0.048 & 0.073 \\
5  & 3.67 & 0.666 & 0.077 & 0.116 \\
10 & 3.70 & 0.674 & 0.130 & 0.193 \\
20 & 3.73 & 0.682 & 0.246 & 0.361 \\
30 & 3.76 & 0.689 & 0.398 & 0.579 \\
\bottomrule
\end{tabular}
\caption{The best-of-$k$ allocation and its truthful price: $\bar f_k$ is the expected maximum evaluator score over $k$ draws; $x(k)=(\bar f_k-1)/4$ its unit-interval transform; $p(k)$ the payment; $\pi(k)=p(k)/x(k)$ the per-click price.}
\label{tab:pricing}
\end{table}

The allocation rises gently whilst the payment, which integrates it, rises the more visibly. This gentleness is itself informative: the steepness of a best-of-$k$ allocation is governed by the diversity of the candidate pool as the evaluator judges it; the present pool is relatively homogeneous, so further optimisation effort yields diminishing return, a fact the mechanism prices faithfully.

\textbf{Non-Monotone Allocations: Ironing and $\varepsilon$-IC.}
The worked example rests upon an allocation monotone by construction; a learned allocation need not be. When the platform prices the responses of a given policy, the composition of policy and evaluator is not constrained to be monotone in the bid: pushed towards ever more prominent integration, a policy may produce responses the evaluator judges less likely to draw a click, since it penalises over-emphasis (Section~\ref{subsec:perturbations}). Two routes recover a principled price: ironing replaces $x$ with the monotone allocation induced by the convex envelope of $U_0$ and prices it by the identity, the ironed mechanism being TIE (Theorem~A.8); alternatively the original $x$ is retained and charged the tight $\varepsilon$-incentive-compatible price, the least achievable slack $\varepsilon^{\star}$ characterised exactly (Theorems~A.9 and~A.10). The framework thus prices any measurable allocation: the platform need neither constrain nor retrain its policy, and the evaluator supplies the click-probability primitive the mechanism-design literature has assumed but not built.

%% file: docs/6-discussion.tex
\section{Discussion and Limitations}
\label{sec:discussion}

We calibrate our claims by delineating the boundaries of the current methodology.

\textbf{Agent Labels Pending Large-Scale Human Validation.}
Our central open limitation is that the agent-grounded click-intent labels have not yet been validated against a large, independent human standard at scale.
The construction is principled and reproducible, and the pairwise study (Section~\ref{subsec:human_validation}) provides initial evidence that the evaluator's orderings are judged sensibly---when its confidence gap exceeds $1.5$, agreement with majority human preference reaches 96\%. However, a broader, multiply-annotated study is required to establish how well the labels track human preference in general.

\textbf{The Gap Between Simulation and Real-World CTR.} 
The evaluator outputs an expected intent score on a continuous 1--5 scale: an ordinal surrogate for engagement, not a calibrated probability (e.g., a 2.5\% CTR). Because click logs do not yet exist for this setting, mapping these scores to real-world click-through rates requires post hoc calibration against live A/B data. The model supplies the ordinal ranking and continuous gradients that optimisation requires; the final scalar projection awaits live data.

\textbf{Relevance-Dominated Label Construction.} 
By design, the relevance gate encodes a strict prior: an off-topic advertisement (dog food pitched to a laptop query) will not be clicked, however well written. The evaluator consequently tracks semantic relevance and text quality first, with the psychological features exerting a bounded modulation---grounding the signal in user utility rather than rewarding manipulative but irrelevant copy.

\textbf{Personality Priors as Structural Boundaries.}
The Big Five trait mapping is grounded in personality-and-persuasion research~\cite{hirsh2012,matz2017}, but the evidentiary support is uneven across traits---the Openness, Extraversion, and Agreeableness pairings are the better attested~\cite{stysko2024}---and recent meta-analytic work counsels caution about the size and robustness of trait-targeted advertising effects in general~\cite{perla2026}. The personality model should therefore be read as a principled, interpretable source of inter-reader variation rather than a calibrated causal model of how traits drive clicks; because the raw objective features are stored independently, the signed weights $s_{t,f}$ and strength coefficient $\lambda$ can be re-tuned post hoc without re-running the simulation.

\textbf{Paths to a Richer Intent Signal.}
The label construction reads the user query through a single channel, topical relevance, which saturates once an advertisement is on-topic; the stage of purchase intent expressed in the query and the profile of the reader beyond sampled personality traits are therefore invisible to the present signal. Both are addressable within the existing architecture: a query-stage feature entering the combination rule alongside the relevance gate, trained on corpora augmented with queries spanning the buying process, and richer reader profiles where platform context supplies them. Live engagement data, once it exists, would both calibrate the ordinal scale and validate these extensions directly. These limitations bound what the signal represents without undermining its use: the direct head still supplies the deterministic gradients that generator training and pricing require, and establishing that predicted intent corresponds to real-world engagement remains the most important direction for future work.

%% file: docs/7-conclusion.tex
\section{Conclusion}
\label{sec:conclusion}

Latent click-through intent is the gating problem for advertising within LLM-generated responses. We have addressed it end to end: constructing reproducible intent supervision through psychologically grounded agent simulation where neither behavioural logs nor calibratable human annotation exist; distilling it into a parameter-efficient, behaviourally validated evaluator whose sensitivity exceeds frontier zero-shot judges and survives transfer to wholly fictional products; and carrying its continuous estimate through to a truthful payment rule, demonstrated on a best-of-$k$ allocation, with ironing and $\varepsilon$-incentive-compatible extensions for the non-monotone case. Two steps remain for practice: calibrating the ordinal intent scale against live engagement data, and training generation policies directly against the differentiable signal. Though instantiated here for LLM answer engines, nothing in the construction is specific to them: the same agent-grounded supervision, ordinal evaluator, and payment identity apply wherever advertisements are embedded within AI-generated responses. We hope this work supplies the missing measurement layer on which a principled economics of LLM-native advertising can be built.